# Decision-making for Autonomous Vehicles on Highway: Deep Reinforcement Learning with Continuous Action Horizon

Hao Chen[1,2], Xiaolin Tang[1] ， Teng Liu[1]

1. College of Mechanical and Vehicle Engineering, Chongqing University, Chongqing 400044, China;

2. Department of Industrial Design, Chongqing Normal University, Chongqing 401331, China

***Abstract*—Decision-making strategy for autonomous vehicles delineating a sequence of driving maneuvers aimed at accomplishing specific navigational missions. In this research, the deep reinforcement learning (DRL) methodology is employed to address the intricate and continuous-horizon decision-making challenge posed by highway scenarios. Initially, the paper introduces the vehicle kinematics and the driving scenario on the freeway. The overarching goal of the autonomous vehicle is to execute a collision-free, efficient, and seamless policy. Subsequently, the specific algorithm, known as proximal policy optimization (PPO)-enhanced DRL, is expounded upon. This algorithm is designed to surmount challenges relating to delayed training efficiency and sample inefficiency, thereby achieving enhanced learning efficiency and exceptional control performance. Finally, the decision-making strategy rooted in PPO-DRL is assessed from multifaceted perspectives, encompassing optimality, learning efficiency, and adaptability. Furthermore, its potential for online application is explored through its utilization in analogous driving scenarios.**

## NOMENCLATURE

| | |
|---|---|
| DRL | Deep Reinforcement Learning |
| PPO | Proximal Policy Optimization |
| AI | Artificial Intelligence |
| AD | Autonomous Driving |
| MDP | Markov Decision Process |
| NN | Neural Network |
| RL | Reinforcement Learning |
| DQL | Deep Q-learning |
| AEV | Autonomous Ego Vehicle |
| IDM | Intelligent Driver Model |
| MOBIL | Minimize Overall Braking Induced by Lane Changes |
| ACC | Adaptive Cruise Control |
| SGD | Stochastic Gradient Descent |
| CEM | Cross-Entropy Method |

## I. INTRODUCTION

MOTIVATED by advancements in artificial intelligence (AI) technologies, autonomous vehicles are emerging as promising transportation solutions aimed at mitigating traffic accidents and enhancing road efficiency [1]-[2]. Four pivotal modules constitute essential components for an automated vehicle, which are perception, decision-making, planning, and control [3]-[4]. Attaining comprehensive automation within intricate driving scenarios necessitates further dedication and exploration across these research domains.

Decision-making encompasses a continuous sequence of driving maneuvers aimed at accomplishing specific navigational tasks [5]-[6]. The distinctive directives within a decision-making strategy typically involve adjustments to the accelerator pedal and steering angle. Numerous endeavors have been undertaken to formulate an appropriate decision-making policy. For instance, Hoel et al. [7] employed a Monte Carlo tree search to derive tactical decision-making strategies for autonomous driving (AD). The driving environment adheres to a partially observable Markov decision process (MDP), and the ensuing outcomes are juxtaposed with a neural network (NN) policy. The authors explored cooperative lane-changing decisions as a means to optimize limited road resources and mitigate competition [8]. Furthermore, Ref. [9] expounded upon highway-exit decisions for autonomous vehicles. The authors asserted that the presented decision-making controller yields an elevated likelihood of successful highway exits, backed by 6000 instances of stochastic simulations.

Reinforcement learning (RL), particularly deep reinforcement learning (DRL) methods, possess significant potential for addressing decision-making challenges in autonomous driving (AD) [10]. For instance, researchers in [11] employed deep Q-learning (DQL) to address the lane-changing decision-making issue within an uncertain highway environment. Similarly, in the context of lane changing, Zhang et al. [12] devised a model-based exploration policy grounded in intrinsic surprise rewards.

This work was supported by the Education Planning Project of Chongqing （No. K22YG205145）(*Corresponding authors: Xiaolin Tang*)

Furthermore, Reference. [13] conducted a comprehensive survey of the prevailing applications of RL and DRL for automated vehicles, encompassing agent training, evaluation techniques, and robust estimation. Nevertheless, several limitations curtail the real-world viability of DRL-based decision-making strategies. These encompass challenges related to sample efficiency, slow learning rates, and operational safety.

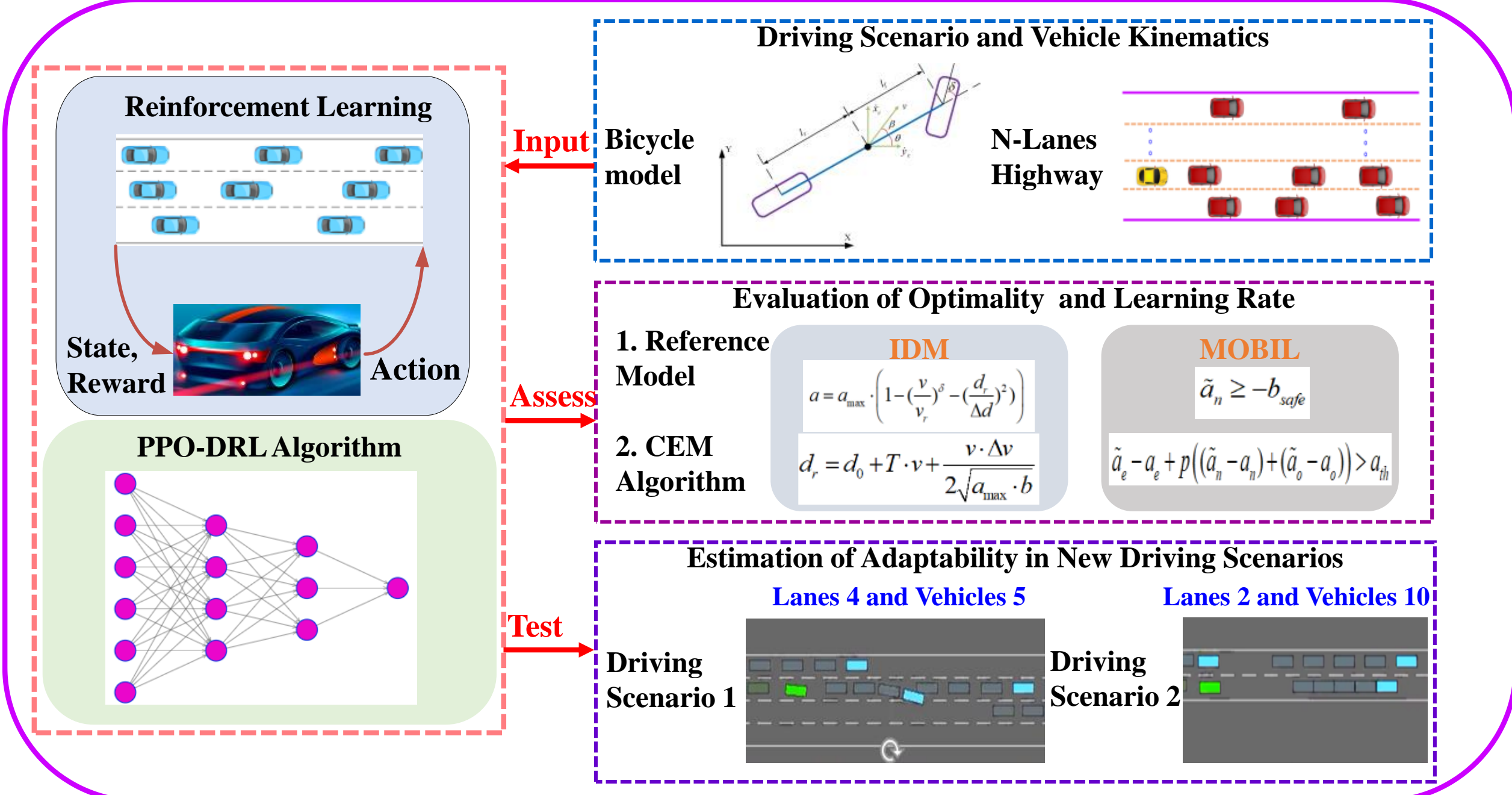

Fig. 1. An efficient and safe decision-making control framework based on PPO-DRL for autonomous vehicles.

This study aims to develop an effective and secure decision-making policy for autonomous driving (AD). To achieve this goal, a proximal policy optimization (PPO)-enhanced deep reinforcement learning (DRL) approach is presented for highway scenarios with a continuous action horizon, as illustrated in Fig. 1. Initially, the vehicle's kinematics and driving scenarios are established, wherein the autonomous ego vehicle is designed to operate efficiently and safely. Through the utilization of the policy gradient method, the PPO-enhanced DRL framework enables direct acquisition of control actions, while maintaining a trust region with bounded objectives. The specific implementation details of this DRL algorithm are subsequently elaborated upon. Ultimately, a series of comprehensive test experiments are designed to assess the optimality, learning efficiency, and adaptability of the proposed decision-making policy within the context of highway scenarios.

This work introduces three key contributions and innovations: 1) the development of an advanced, efficient, and safe decision-making policy for AD on highways; 2) the application of PPO-enhanced DRL to address the transferred control optimization challenge in autonomous vehicle scenarios; 3) the establishment of an adaptive estimation framework to assess the adaptability of the proposed approach. This endeavor represents a concerted effort to enhance the efficiency and safety of decision-making policies through the utilization of cutting-edge advanced DRL methodologies.

To elucidate the contributions of this article, the subsequent sections are organized as follows. Section II outlines the vehicle kinematics and driving scenarios on the highway. The PPO-enhanced DRL framework employed in this research is expounded upon in Section III. The ensuing section, Section IV, delves into the analysis of pertinent simulation outcomes for the proposed decision-making strategy. Ultimately, Section V presents the concluding remarks.

## II. Vehicle Kinematics and Driving Scenarios

In this section, we establish the highway driving scenario for our research. This environment encompasses the autonomous ego vehicle (AEV) and the surrounding vehicles. We also detail the vehicle kinematics of these entities, allowing for the calculation of longitudinal and lateral speeds. Additionally, we introduce reference models for driving maneuvers in both the longitudinal and lateral directions.

### A. *Vehicle Kinematics*

In this study, we elucidate the vehicle kinematics through the application of the widely acknowledged common bicycle model [14]-[15], characterized by nonlinear continuous horizon equations. The representation of the inertial frame is illustrated in Fig. 2. Computation of the differentials for position and inertial heading is outlined as follows:

$$\dot{x} = v\cos(\psi + \beta) \tag{1}$$

$$\dot{y} = v\sin(\psi + \beta) \tag{2}$$

$$\dot{\psi} = \frac{v}{l_r}\sin\beta \tag{3}$$

where $(x, y)$ represents the positional coordinates of the vehicle within the inertial frame. The vehicle velocity is denoted as $v$, while $l_r$ signifies the distance between the center of mass and the rear axles. Additionally, $\psi$ represents the inertial heading, and $\beta$ denotes the slip angle at the center of gravity. This angle and the vehicle speed can be further displayed as:

$$\dot{v}=a \tag{4}$$

$$\beta=\arctan\left(\frac{l_r}{l_f+l_r}\tan(\delta_f)\right) \tag{5}$$

where $l_f$ represents the distance of the center of mass from the front. $\delta_f$ signifies the front steering angle. The two degrees-of-freedom model simplifies the delegation of the primary vehicle parameters: speed and acceleration. In this model, the control inputs encompass acceleration and steering angle, forming a continuous-time horizon as discussed in this article.

The default parameters for both the AEV and the surrounding vehicles are identical. The dimensions consist of a length of 5.0 m and a width of 2.0 m. The initial speed is selected randomly from the range of [23, 25] m/s, while the maximum achievable speed is limited to 30 m/s. The initial position is stochastically assigned along the highway, thus reflecting the inherent uncertainty of the driving environment.

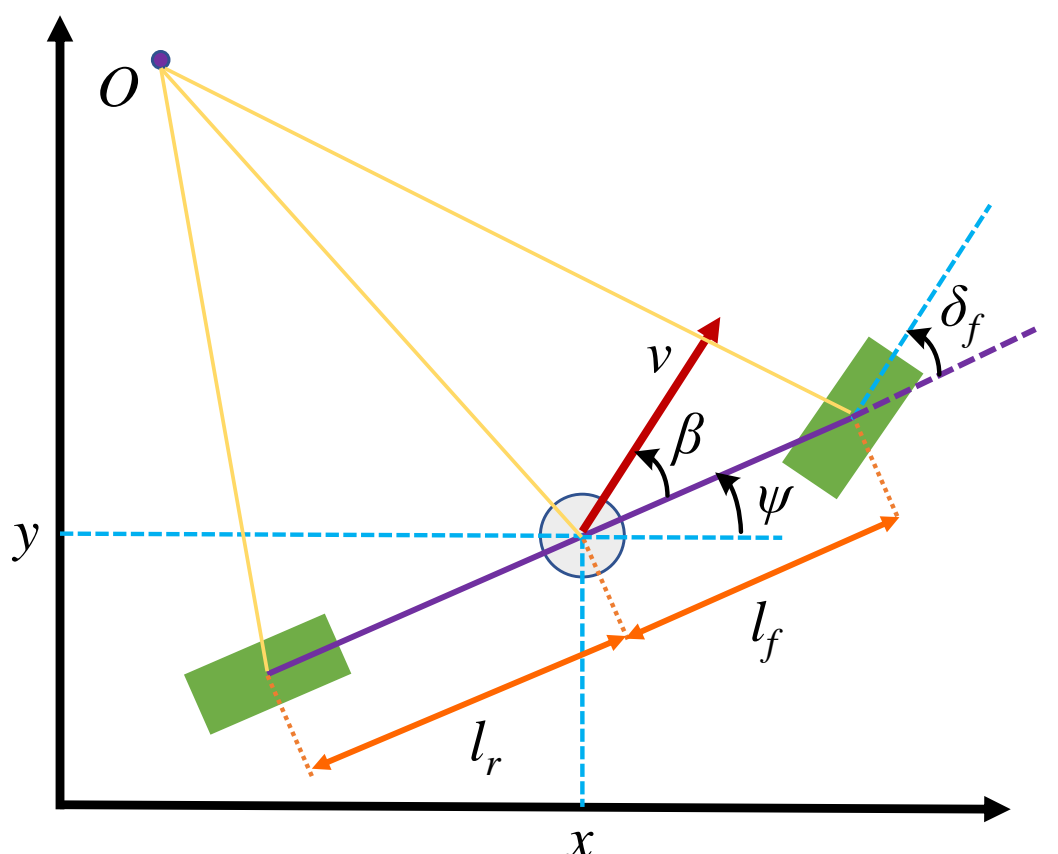

Fig. 2. Bicycle model for vehicle kinematics on highway.

## B. Driving Scenarios

To emulate a real-world driving environment on the highway, we establish a driving scenario featuring $N$ lanes in the same direction, as illustrated in Fig. 3. The decision-making strategy for the AEV presented in this study involves ascertaining control actions for vehicle speed and steering angle at each time step. The primary aim of the AEV is to achieve maximum speed while avoiding collisions with surrounding vehicles.

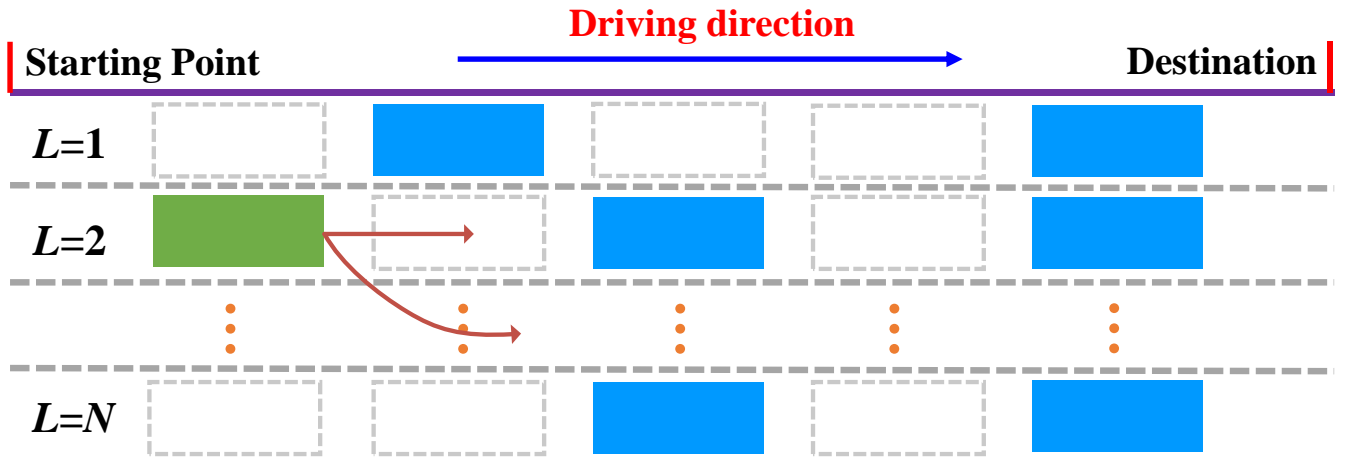

Fig. 3. Driving scenario on highway with $N$ lanes for decision-making policy.

The overtaking behavior signifies instances when the subject vehicle surpasses nearby vehicles through a combination of lane-changing and accelerating maneuvers. Typically, the evaluation of decision-making policies hinges on three key indicators: safety, efficiency, and comfort. Safety necessitates that the AEV must steer clear of collisions. Efficiency entails the preference of an autonomous vehicle to elevate its speed. Comfort, on the other hand, signifies that the AEV should manage the frequency of lane-changing and the degree of vehicle deceleration [16].

In this study, the primary considerations for the AEV encompass safety and efficiency. Additionally, the vehicle exhibits a preference for positioning itself within the high-speed lane. As illustrated in Fig. 3, the AEV is represented by the green vehicle, while the surrounding vehicles are depicted in blue. In each lane, the total count of surrounding vehicles is denoted as $K$. Within this article, an episode is defined as the scenario where the AEV successfully overtakes all the surrounding vehicles or reaches its designated destination.

Without sacrificing generality, we designate the number of lanes on the highway as $N$=3. Within each lane, the count of surrounding vehicles is configured to $K$=5. The Autonomous Ego Vehicle (AEV) adheres to a predetermined lane, specifically the right lane. The simulation operates at a frequency of 20 Hz, with a sampling time of 1 second (implying that the AEV makes decisions every second). An episode extends for a duration of 50 seconds. The driving behavior of the surrounding vehicles adheres to two commonly utilized models, which will be elaborated upon in the ensuing subsection.

## C. Behavioral Controller

In this section, we formulate the Intelligent Driver Model (IDM) and the Minimize Overall Braking Induced by lane changes (MOBIL) to govern the driving behaviors of surrounding vehicles. Furthermore, the fusion of these two models serves as a benchmark for the AEV, facilitating a comparison with the proposed DRL approach.

The IDM is commonly employed for Adaptive Cruise Control (ACC) in automated vehicles, functioning as a continuous-time horizon car-following model [17]-[18]. The longitudinal acceleration within IDM is expressed as follows:

$$a=a_{\max}\cdot\left(1-(\frac{v}{v_r})^{\delta}-(\frac{d_r}{\Delta d})^2)\right) \tag{6}$$

where $a_{\max}$ is the maximum acceleration. $v_r$ and $d_r$ represent the requested vehicle velocity and separation distance, respectively. $\delta$ stands as the constant acceleration parameter, while $\triangle d$ pertains to the interval between the subject vehicle and the leading vehicle. Within the framework of the IDM, the requested speed is determined by the interplay between the maximum acceleration and the requested distance, with the latter calculated as follows:

$$d_r=d_0+T\cdot v+\frac{v\cdot\Delta v}{2\sqrt{a_{\max}\cdot b}} \tag{7}$$

where $d_0$ represent the minimum relative distance between two vehicles on the same lane, while $T$ signifies the desired time interval for ensuring safety. $\triangle v$ accounts for the relative speed gap between the subject vehicle and the vehicle ahead, and $b$ denotes the value of deceleration aligned with comfortable criteria. The specific parameters of the IDM employed in this study are detailed in Table I.

TABLE I
DEFAULT PARAMETERS OF IDM

| Symbol | Value | Unit |
|---|---|---|
| Maximum acceleration $\boldsymbol{a_{max}}$ | 6 | m/s$^2$ |
| Acceleration argument $\boldsymbol{\delta}$ | 4 | / |
| Desired time gap $\boldsymbol{T}$ | 1.5 | s |
| Comfortable deceleration rate $\boldsymbol{b}$ | -5 | m/s$^2$ |
| Minimum relative distance $\boldsymbol{d_0}$ | 10 | m |

Upon establishing the longitudinal acceleration of the adjacent vehicle, MOBIL is employed to govern lateral lane-changing decisions [19]. The MOBIL framework incorporates two pivotal conditions: the safety criterion and the incentive condition. The safety criterion dictates that during a lane change, the subsequent vehicle should avoid excessive deceleration to prevent collisions. The mathematical representation of this acceleration constraint is as follows:

$$\tilde{a}_n \geq -b_{safe} \tag{8}$$

where $\tilde{a}_n$ denotes the acceleration experienced by the newly following vehicle after executing a lane change, while $b_{safe}$ represents the upper limit for deceleration applied to the new follower. Equation (8) is strategically employed to establish conditions that guarantee collision-free scenarios.

Assuming the $a_n$ and $\tilde{a}_n$ denote the accelerations of the new follower prior to and after lane-changing, while $a_o$ and $\tilde{a}_o$ refer to the accelerations of the previous follower before and after the lane-change event. The incentive condition is established through the imposition of an acceleration constraint, which is articulated as follows:

$$\tilde{a}_e - a_e + p\left((\tilde{a}_n - a_n) + (\tilde{a}_o - a_o)\right) > a_{th} \tag{9}$$

where $a_e$ and $\tilde{a}_e$ represent the accelerations of the AEV prior to and following a lane change. $p$ corresponds to the politeness coefficient, which quantifies the degree of influence exerted by the followers during the lane-changing process. Additionally, $a_{th}$ denotes the threshold for making lane-changing decisions. This criterion signifies that the intended lane must offer a higher level of safety compared to the current one. Notably, the accelerations employed in the MOBIL approach are determined by the IDM at each time step. Moreover, the AEV possesses the capability to execute overtaking maneuvers by transitioning between the right and left lanes. The specific parameters for MOBIL are outlined in Table II.

TABLE II
MOBIL CONFIGURATION

| Keyword | Value | Unit |
|---|---|---|
| Safe deceleration limitation $\boldsymbol{b_{safe}}$ | 2 | m/s$^2$ |
| Politeness factor $\boldsymbol{p}$ | 0.001 | / |
| Lane-changing decision threshold $\boldsymbol{a_{th}}$ | 0.2 | m/s$^2$ |

## III. PROXIMAL POLICY OPTIMIZATION-ENABLED DEEP REINFORCEMENT LEARNING

This section elucidates the procedural steps involved in implementing the PPO-enhanced DRL method under study. Initially, the foundations of reinforcement learning (RL) methods and the rationale behind employing a continuous-time horizon are presented. Subsequently, the conventional format of the policy gradient technique is detailed. Finally, the utilization of PPO-enabled DRL is illuminated, facilitating the derivation of a decision-making strategy for the control problem established in Section II.

### A. Necessity of Continuous Horizon

RL has emerged as a methodological approach to tackle sequential decision-making problems through a trial and error process [20]-[22]. This dynamic process is exemplified by the interplay between an intelligent agent and its environment. The agent takes control actions within the environment and subsequently receives evaluations of its choices from the environment [23]-[24]. Broadly, RL methods are categorized into policy-based approaches (i.e., policy gradients algorithm) and value-based approaches (i.e., Q-learning and Sarsa algorithms).

In the context of the highway decision-making problem, the intelligent agent functions as the decision-making controller for the AEV, while the surrounding vehicles comprise the environment. This interaction is typically emulated through the application of Markov decision processes (MDPs) with Markov property [25]. The MDP is characterized by a pivotal tuple ($\boldsymbol{S}$, $\boldsymbol{A}$, $\boldsymbol{P}$, $\boldsymbol{R}$, $\gamma$), where $\boldsymbol{S}$ and $\boldsymbol{A}$ are the sets of state variable and control actions. $\boldsymbol{P}$ denotes the transition model of the state variable, and $\boldsymbol{R}$ corresponds to the reward model associated with the state-action pair ($s$, $a$). $\gamma$ is referred to as the discount factor, serving to strike a balance between immediate and future rewards.

The goal of RL techniques is to select a sequence of control actions from set $\boldsymbol{A}$ to maximize cumulative rewards. The cumulative rewards, denoted as $R_t$ is the sum of the current reward and the discounted future rewards as:

$$R_t = \sum_{t=0}^{\infty} \gamma^t \cdot r_t \tag{10}$$

where $t$ is the time step, and $r_t$ represents the corresponding instantaneous reward. Two distinct value functions are defined to convey the significance of selecting control actions. These functions are identified as the state-value function $V$ and the state-action function $Q$:

$$V^{\pi}(s_t) \doteq E_{\pi}[R_t \mid s_t, \pi] \tag{11}$$

$$Q^{\pi}(s_t, a_t) \doteq E_{\pi}[R_t \mid s_t, a_t, \pi] \tag{12}$$

where $\pi$ is a special control policy. It is evident that various control policies result in distinct values for the value function, with the pursuit of optimal performance being desirable. The optimal control policy is defined as follows:

$$\pi(s_t) = \arg\max_{a_t} Q(s_t, a_t) \tag{13}$$

The essence of the RL algorithms lies in updating value functions based on the interactions between the agent and the environment [26]. These value functions subsequently guide the agent in discovering an optimal control strategy. For DRL, the value function is approximated using a neural network. From (12), the state-action function assumes the form of a matrix, with its rows and columns corresponding to the quantities of state variables and control actions. In scenarios characterized by extensive state variable and control action spaces, the process of updating the value function and searching for an appropriate control policy can become inefficient.

To address this limitation, this study models the control actions as the vehicle's throttle and steering angle. The throttle governs acceleration, while the steering angle directly impacts lane-changing behavior. These two actions operate within continuous-time horizons, specifically within ranges of [-5, 5] m/s$^2$ for acceleration and [-Π/4, Π/4] rad (Π is the circumference as 3.1415) for steering angle. This approach enables the AEV to iteratively determine control action pairs at each time step, thereby establishing the vehicle's kinematics as detailed in Section II.A.

### *B. Policy Gradient*

In policy-based RL methods, an estimator of the policy gradient is computed for a stochastic policy, which is represented as follows:

$$\hat{g}=\hat{\mathrm{E}}_t\left[\nabla_\theta \log \pi_\theta(a_t \mid s_t)\hat{A}_t\right] \tag{14}$$

where $\hat{E}_t$ indicates the expectation over a finite batch of samples, $\pi_\theta$ denotes a random control policy, and $\hat{A}_t$ signifies the advantage function. In order to compute the policy gradient estimator, the loss function for updating an RL policy is described as follows:

$$L^{PG}(\theta)=\hat{\mathrm{E}}_t\left[\log \pi_\theta(a_t \mid s_t)\hat{A}_t\right] \tag{15}$$

In the common policy gradient method, this loss function $L^{PG}$ is employed to undergo multiple optimization steps using the same control policy. Nevertheless, challenges can arise during the updating of extensive policies, including issues of sample inefficiency, policy diversity, and hesitance in exploration and exploitation. In order to tackle these challenges, the PPO algorithm is proposed in [27] aiming to merge the strengths of conventional value-based and policy-based RL approaches..

### *C. PPO DRL*

In the traditional policy gradient approach, the policy can undergo rapid changes with each update. In order to mitigate this issue, a policy surrogate objective is adjusted according to the following formulation:

$$L^{CLIP}(\theta)=\hat{\mathrm{E}}_t\left[\min(r_t(\theta)\hat{A}_t, clip(r_t(\theta),1-\tau,1+\tau))\hat{A}_t\right] \tag{16}$$

where

$$r_t(\theta)=\frac{\pi_\theta(a_t \mid s_t)}{\pi_{\theta_{old}}(a_t \mid s_t)} \tag{17}$$

where $r_t(\theta)$ represents the probability ratio. Equation (16) compares two terms: the first term stands as the surrogate objective [28], while the second term refines this surrogate objective by applying a clipping mechanism to the probability ratio. Here, τ s a hyperparameter set to a value of 0.2. In the second term, the probability ratio $r_t(\theta)$ is bounded within the range of 1- τ to 1+ τ, subsequently forming the clipped objective through multiplication with the advantage function. The inclusion of this clipped version serves to prevent excessively significant updates to the policy derived from the previous policy.

In order to establish parameter sharing between the policy and value functions using a neural network, the loss function is reformulated by combining the policy surrogate with an error term from the value function [27]. The resulting modified loss function is constructed as follows:

$$L^{CLIP+VF+S}(\theta)=\hat{\mathrm{E}}_t\left[L_t^{CLIP}(\theta)-c_1L_t^{VF}(\theta)-c_2S[\pi_\theta](s_t)\right] \tag{18}$$

where $L_t^{VF}(\theta)$ is the squared-error loss of the state-value function $(V_\theta(s_t)-V_t^{tar})^2$, $S$ indicates an entropy loss. $c_1$ and $c_2$ are the coefficients.

TABLE III
IMPLEMENTATION CODE OF PPO ALGORITHM

| **PPO Algorithm, Actor-Critic Style** |
|---|
| **1. For** iteration = 1, 2, …, **do** |
| **2.** **For** actor = 1, 2, …, $M$ **do** |
| **3.** Run policy $\pi_{\theta_{old}}$ in environment for $T$ timesteps |
| **4.** Calculate advantage function based on (19)-(20), $\hat{A}_1,\ldots,\hat{A}_T$ |
| **5.** **end for** |
| **6.** Optimize loss function in (18) with respect to $\theta$ for Z epochs |
| **7.** Update $\theta_{old}$ with $\theta$ |
| **8. end for** |

To estimate the advantage function, data from $T$ timesteps ($T$ is much less than the episode length) is sampled. This gathered data is employed to update the loss function, wherein the advantage function is incorporated in a truncated form as follows:

$$\hat{A}_t=\delta_t+(\gamma\lambda)\delta_{t+1}+\ldots+(\gamma\lambda)^{T-t+1}\delta_{T-1} \tag{19}$$

where

$$\delta_t=r_t+\gamma V(s_{t+1})-V(s_t) \tag{20}$$

During each iteration, $M$ actors are established to gather data over $T$ timesteps data. $\lambda$ represents the discounting factor for the advantage function. Subsequently, the surrogate loss is formulated using the collected data and optimized through mini-batch stochastic gradient descent (SGD) for $Z$ epochs. The implementation pseudo-code for the PPO algorithm is presented in Table III.

As explained in Section III.A, the control actions encompass the vehicle throttle and steering angle, both operating within a continuous time horizon. The state variables consist of the AEV's speed, position, as well as the relative speed and position between the AEV and its surrounding vehicles:

$$\Delta s=\left|s_{aev}-s_{sur}\right| \tag{21}$$

$$\Delta v=\left|v_{aev}-v_{sur}\right| \tag{22}$$

where $s$ and $v$ are the position and speed information obtained from (1)-(5) in Section II.A. The superscript *aev* and *sur* indicate the AEV and surrounding vehicles, respectively. The expressions (21) and (22) can also be treated as the transition model $\boldsymbol{P}$ in the RL framework.

Finally, the reward function $\boldsymbol{R}$ in this article encompasses three components, reflecting the objectives of efficiency, safety, and lane preference. Specifically, the AEV aims to maximize its speed, prioritize the right lane, and prevent collisions with other surrounding vehicles. The instantaneous reward at time step $t$ is defined as follows:

$$r_t = -100 \cdot collision - 40*(L\text{-}1)^2 \text{-} 10*(v_{aev} - v_{aev}^{\max})^2 \qquad (23)$$

where $collision \in \{0, 1\}$ indicates the collision conditions for the AEV. $L \in \{1, 2, 3\}$ implies the lane number. For the sake of easy comparison, the value of the instantaneous reward is scaled to the range [0, 1] at each step. This scaling implies that the maximum value of the cumulative rewards for one episode equals the duration time (set as 50 in this work) of the driving scenario.

The default parameters for the presented PPO-enhanced DRL method are defined as follows: The discount factor $\gamma$ and learning rate $\alpha$ in the RL framework are 0.8 and 0.01. The training timesteps $N$ is 51200, the mini-batch size $Z$ is 64, the hyperparameter $\tau$ is 0.2, and the discounting coefficient for advantage function $\lambda$ is 0.92. The decision-making policy on the highway for the AEV is derived and estimated in the OpenAI gym Python toolkit [29]. The performance of this proposed decision-making strategy is discussed and analyzed in the subsequent section.

## IV. Experiments and Discussion

This section evaluates the control performance of the proposed PPO-DRL-based decision-making strategy for the AEV. The evaluation comprises three key aspects. Firstly, the efficacy of this decision-making policy is compared and validated against two alternative methods. The comprehensive simulation results demonstrate its optimality. Secondly, the learning capability of the proposed PPO algorithm is substantiated through an analysis of the loss function and the accumulation of rewards. Lastly, the derived decision-making policy is evaluated in two comparable highway driving scenarios to showcase its adaptability.

### *A. Effectiveness of PPO DRL*

This subsection presents a comparative analysis of three decision-making methods for highway scenarios: PPO-DRL, the reference model (IDM+MOBIL) , and the cross-entropy method (CEM) . Notably, CEM has been demonstrated as effective for addressing continuous-time horizon problems in previous research [30]. The reference model and CEM are considered as benchmark approaches, serving to establish the optimality of the PPO-DRL algorithm. It is important to highlight that the setting parameters in both the PPO and CEM methods remain consistent.

The overall reward accumulated in each episode serves as a key indicator of the performance of the control policy in DRL. The normalized average rewards across the three compared techniques are illustrated in Fig. 4. The upward trajectory of these curves signifies that the AEV progressively improves its performance through interaction with the environment. Furthermore, it is evident that the learning rate exhibited by PPO-DRL surpasses that of the other two methods. The rewards consistently outperform those achieved by CEM and IDM+MOBIL. Consequently, the control policy derived through the PPO approach demonstrates superiority over the other two strategies.

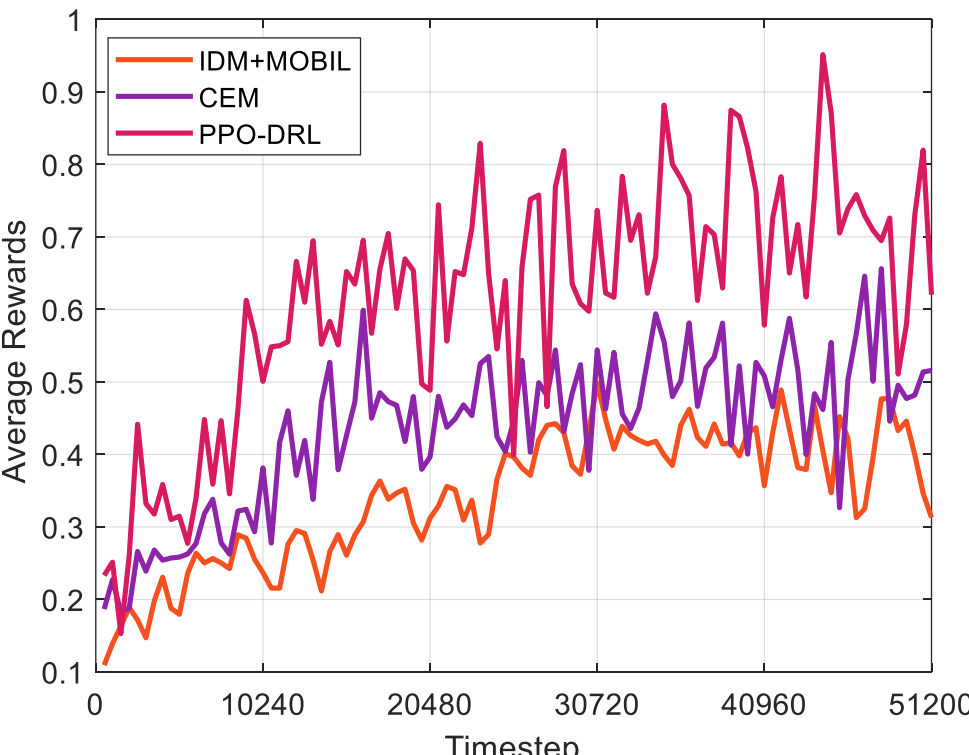

Fig. 4. Normalized average rewards in the reference model, CEM, and PPO-DRL for comparison purposes.

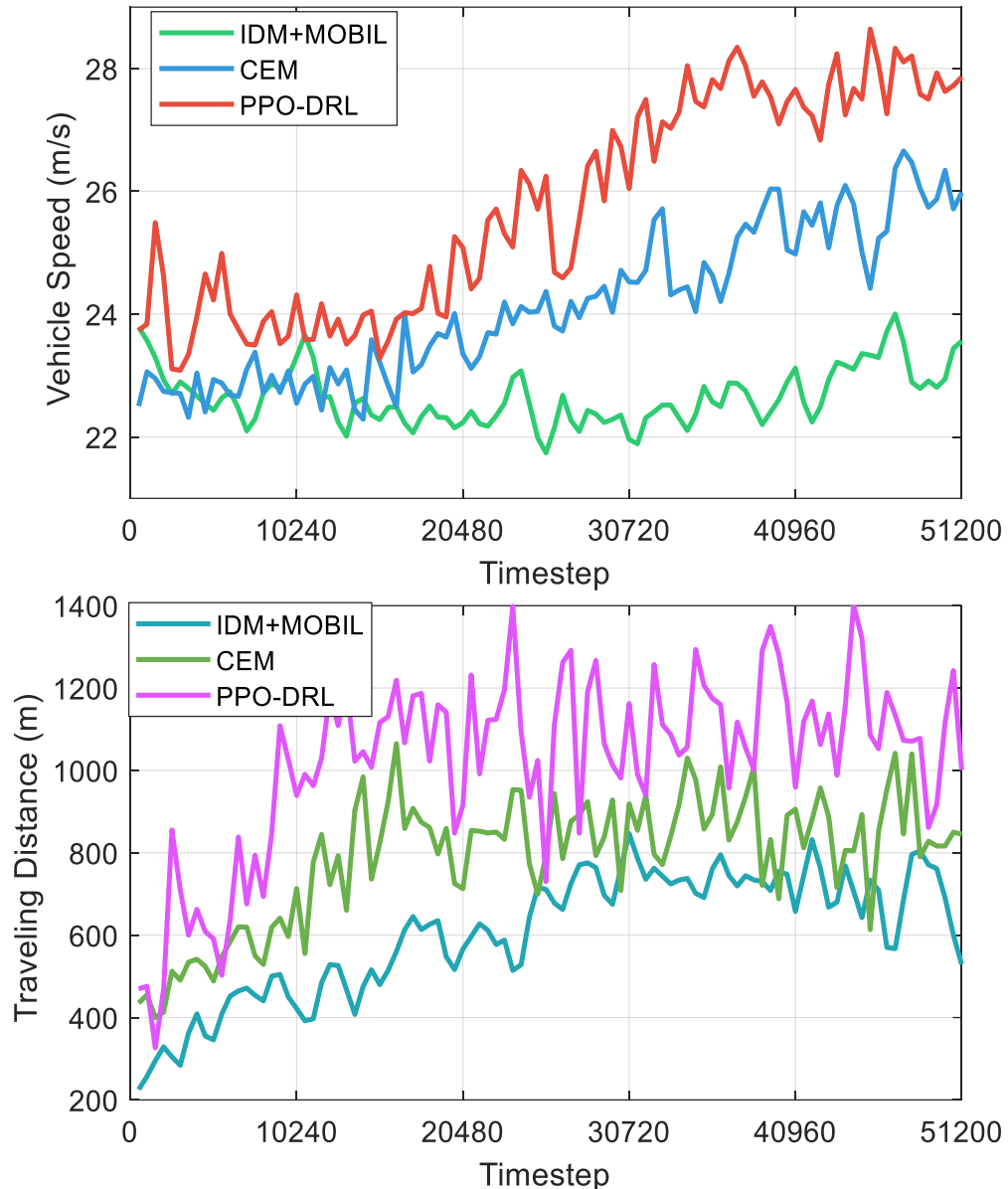

Fig. 5. Vehicle speed and traveling distance of the AEV in three methods.

Given that vehicle speed and distance serve as the selected state variables in this study, Fig. 5 illustrates the diverse trajectories of these variables. In (23), the reward function prompts the AEV to accelerate at suitable instances, thus correlating higher speed with greater rewards. Additionally, the increased distance traveled by the AEV signifies that the chosen control actions facilitate extended travel and collision avoidance. These simulation outcomes underscore the AEV's capacity to attain efficiency and safety objectives effectively under the guidance of the PPO-DRL approach.

Finally, to juxtapose the performance of these three methods within collision-free scenarios, Table IV delineates the collision rate and success rate across testing episodes (the number of testing episodes is 100). The collision rate signifies the likelihood of a collision occurring, while the success rate denotes the AEV's ability to overtake all surrounding vehicles and reach its destination. Evidently, the PPO-DRL approach demonstrates superior collision avoidance capabilities compared to the other two methods. Furthermore, the elevated success rate value underscores the PPO algorithm's proficiency in efficiently accomplishing the driving task.

TABLE IV
COLLISION CONDITIONS IN THREE COMPARED APPROACHES

| Algorithms | Collision rate (%) | Success rate (%) |
| --- | --- | --- |
| PPO-DRL | 0.59 | 99.03 |
| CEM | 4.32 | 91.55 |
| IDM+MOBIL | 7.10 | 87.21 |

## B. *Learning rate of PPO DRL*

In this subsection, we delve into the learning rate and convergence rate of the introduced PPO-DRL approach. The primary goal of DRL algorithms is to update the state-action function $Q(s, a)$ in distinct manners. The loss function elucidated in (18) encapsulates the virtues of a chosen control policy. The aggregate loss for both PPO and CEM is graphically depicted in Fig. 6. The conspicuous downward trajectories signify that the AEV progressively refines its control policy through iterative trial and error. Furthermore, it's noteworthy that the loss values for PPO consistently remain lower than those for CEM. This observation suggests that the AEV governed by the PPO algorithm attains a higher level of familiarity with the driving environment compared to CEM. Thus, it can be confidently asserted that the convergence rate of PPO outperforms that of CEM when tackling the decision-making problem on the highway.

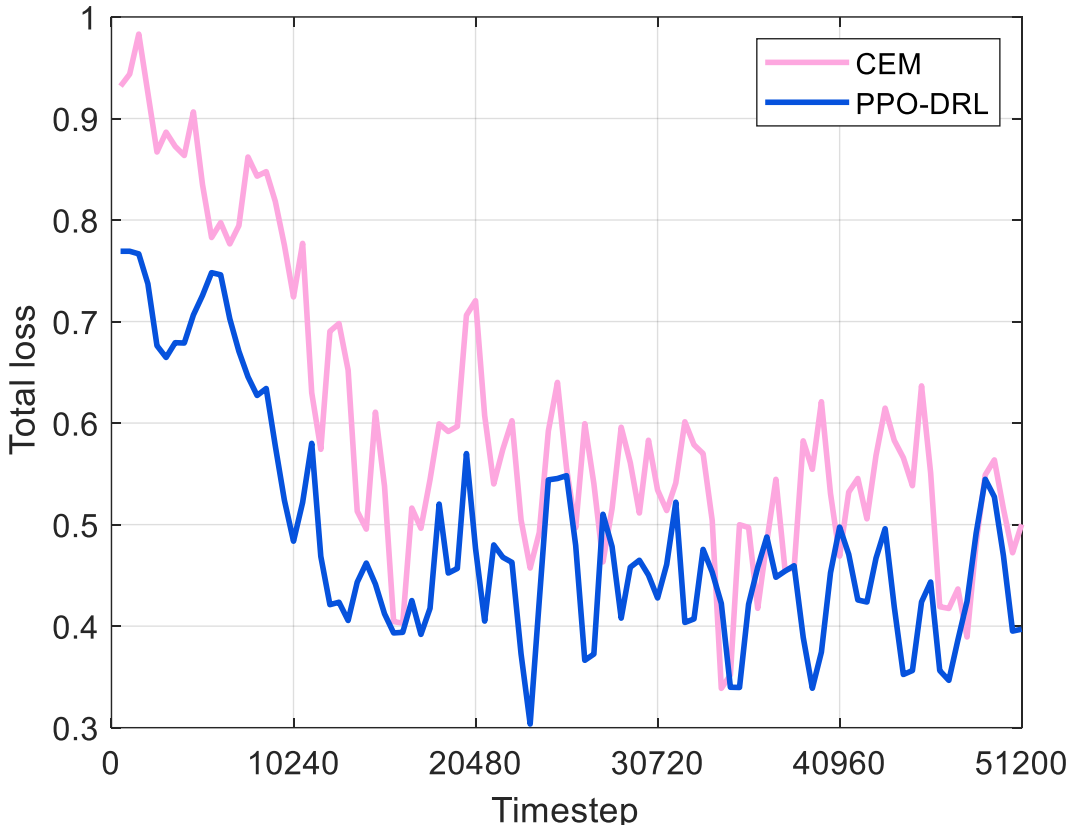

Fig. 6. Value of loss function in two DRL methods: CEM and PPO.

To scrutinize the learning rate disparities between the PPO and CEM algorithms, Fig. 7 illustrates the trajectories of cumulative rewards across these two methodologies. As defined in (10), the cumulative rewards encompass the summation of the current reward and discounted future rewards, thereby serving as a pivotal determinant for control action selection. Evidently portrayed in Fig. 7, the PPO consistently outperforms the CEM, signifying that the control policy furnished by PPO is superior. The AEV operating under the PPO paradigm acquires a more extensive repertoire of knowledge and experiential insights pertaining to the driving environment. This augmentation can be ascribed to the innovative loss function delineated in (18), which expedites the intelligent agent's quest for an optimal control policy.

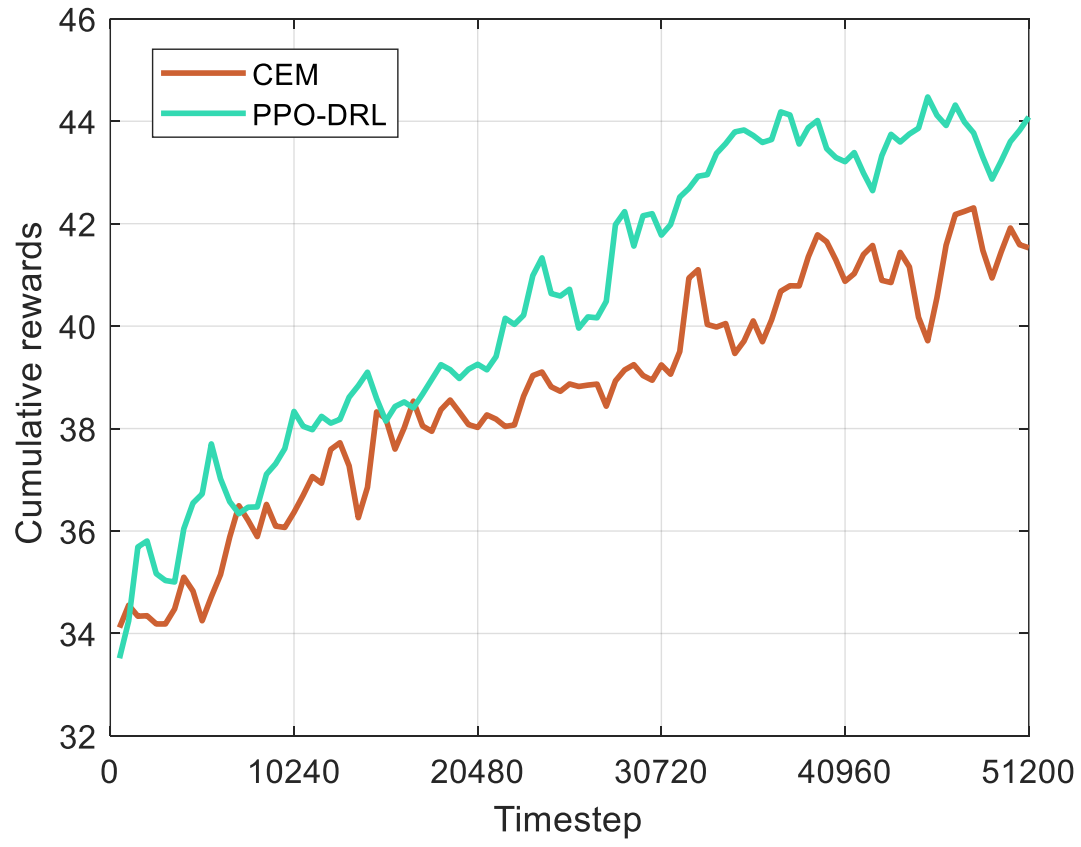

Fig. 7. Accumulated rewards in two compared algorithms: CEM and PPO.

## C. *Adaptability of PPO DRL*

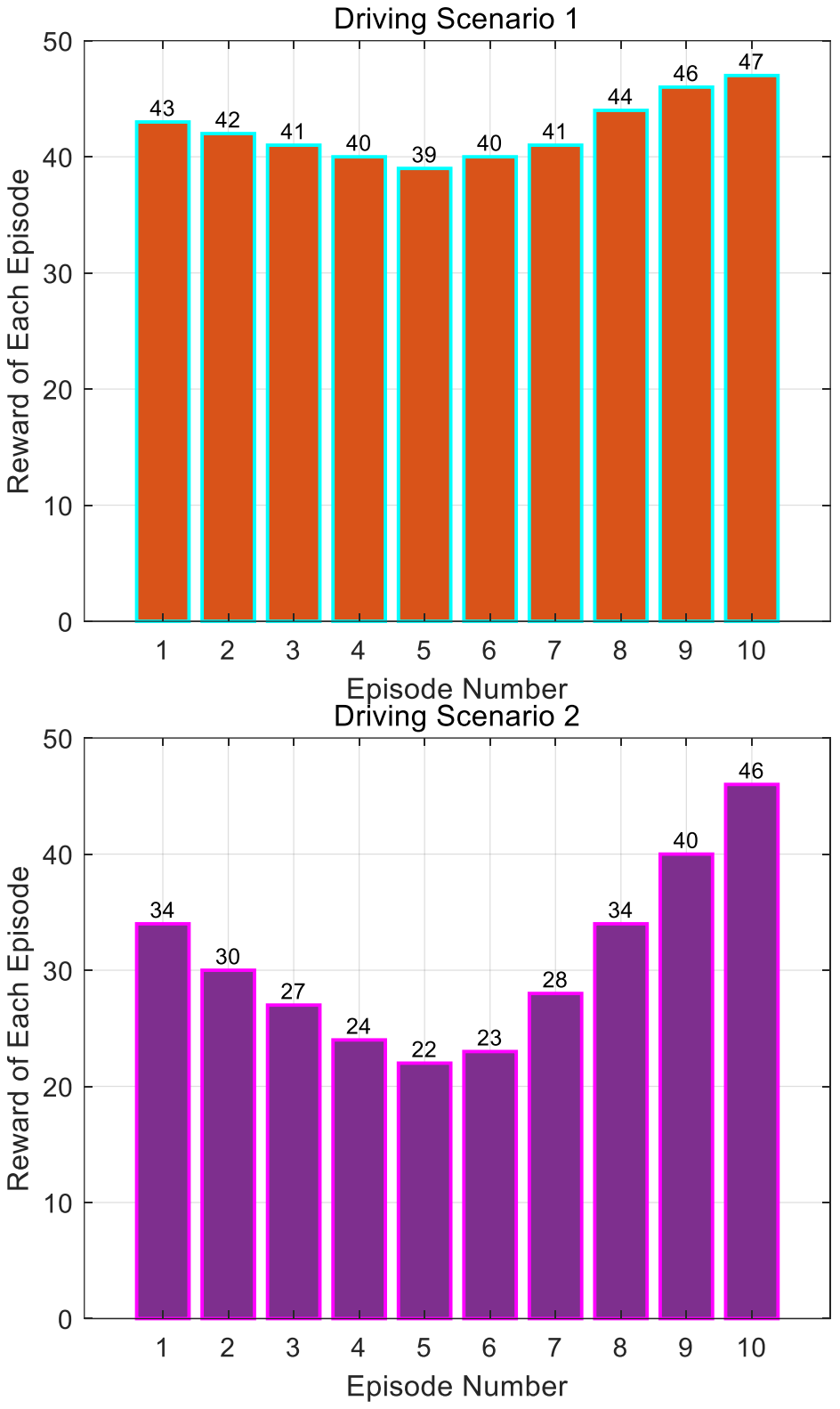

Fig. 8. Rewards in the testing experiments for two new driving scenarios.

This subsection introduces an adaptive estimation framework aimed at substantiating the efficacy of the proposed decision-making policy. The fundamental determinants of a given driving scenario encompass the count of lanes and vehicles therein. In this vein, we manipulate these parameters to instantiate two novel driving scenarios. The first scenario entails four lanes, each accommodating five vehicles (designated as driving scenario 1). Conversely, the second scenario entails two lanes, with ten vehicles occupying each lane (designated as driving scenario 2). A total of 10 testing episodes are conducted for each of these distinct scenarios. The speed and position attributes of these encompassing vehicles are also subject to randomized assignment. These designed driving scenarios serve as emblematic representations of the variegated uncertainties inherent in actual driving environments. Moreover, they facilitate a lucid demonstration of the adaptability inherent in the proposed decision-making policy..

Fig. 8 depicts the aggregate rewards achieved by the PPO algorithm within these two novel driving scenarios. A higher reward value signifies a more fitting control policy tailored to the specific scenario. Conversely, lower reward values can be attributed to two distinct factors. Firstly, the random positioning of the surrounding vehicles may lead to the obstruction of all lanes, impeding the AEV's ability to execute efficient lane changes. Secondly, the AEV might engage in hazardous lane-changing maneuvers in exceptional circumstances, resulting in collisions. Notably, Fig. 8 showcases that the acquired decision-making policy exhibited superior performance within the context of the first scenario. This can be attributed to the fact that, in the initial driving scenario, an additional lane was introduced while maintaining the same count of surrounding vehicles. This augmentation in lane availability provides the AEV with increased opportunities for successful lane-changing and collision avoidance. To further elucidate the adaptability of the proposed decision-making policy, we meticulously analyze two specific episodes.

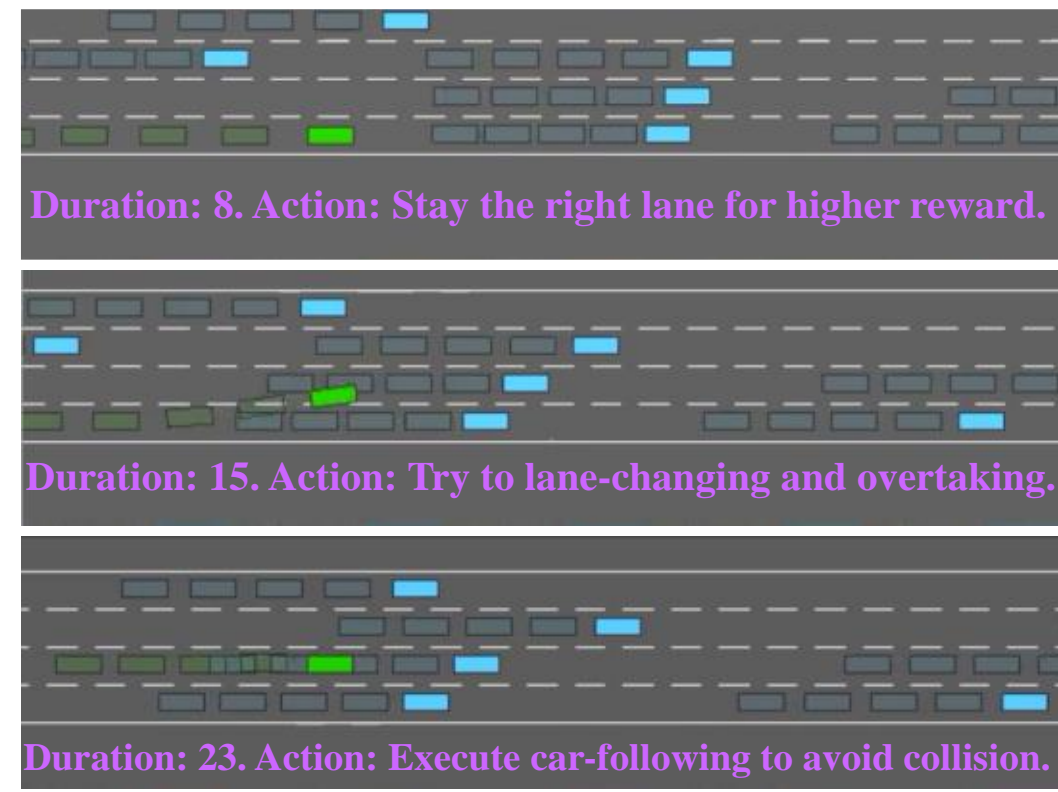

Fig. 9. Episode 5 in driving scenario 1: car-following to avoid collision.

In the context of the new driving scenario 1, episode 5 is selected to analyze due to the lowest value of the reward. As depicted in Fig. 9, all lanes remain obstructed by surrounding vehicles for an extended duration. Consequently, the AEV is compelled to engage in car-following maneuvers to avert potential collisions. This circumstance necessitates a reduction in the AEV's speed, as it patiently awaits an opportune moment to execute an overtaking maneuver. The episode's reward is influenced by factors such as collision occurrences, vehicle velocity, and lane preference, resulting in a slightly diminished value compared to other episodes. This outcome is not unexpected, as safety assumes paramount importance within real-world driving scenarios. This observation underscores the adaptability of the learned decision-making policy, which demonstrates its capacity to adeptly respond to dynamically changing driving conditions.

Fig. 10 provides an illustration of episode 6 within driving scenario 2, utilizing the PPO-DRL-enabled decision-making strategy. Notably, this scenario exhibits a reduction in the number of lanes accompanied by an increase in the number of vehicles. As a result, the AEV faces heightened complexity in decision-making due to the more intricate driving environment. The graphical representation in Fig. 10 reveals that the AEV undertakes a daring lane-changing maneuver in the presence of numerous surrounding vehicles. In the training procedure, the AEV might not have encountered this particular situation, rendering it challenging to accurately predict potential collisions in such circumstances. To address this challenge, two potential research avenues could be pursued to enhance the AEV's decision-making capabilities. Firstly, extending the training process could allow the AEV to accumulate more insights from diverse driving environments, thereby refining its decision-making skills. Secondly, the integration of communication technology to provide the AEV with real-time information about its surroundings could facilitate more informed and judicious decision-making on the highway.

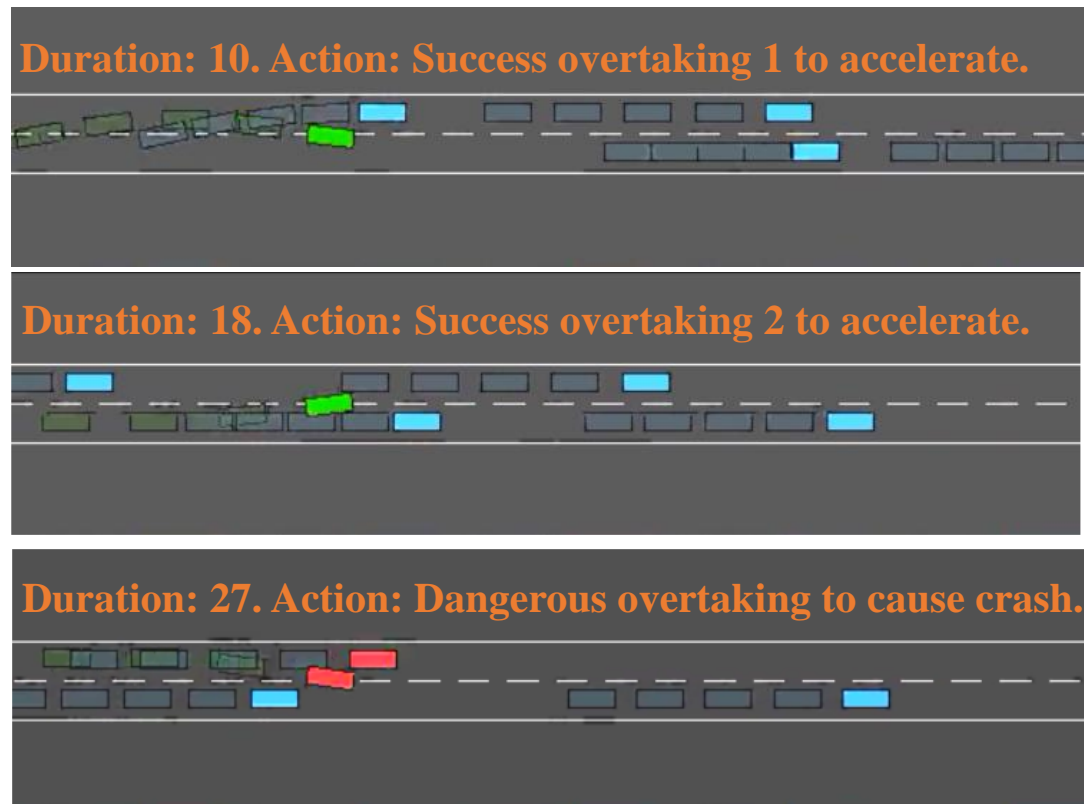

Fig. 10. Episode 6 in driving scenario 2: dangerous overtaking to cause crash.

## V. Conclusion

This study introduces an effective and secure decision-making policy for autonomous vehicles on the highway, employing the PPO-DRL methodology. The developed control framework is applicable to analogous driving scenarios featuring varying lane configurations and surrounding vehicles. The simulation outcomes unequivocally demonstrate the superior performance of the proposed decision-making approach in terms of optimality, convergence rate, and adaptability. Notably, the derived decision-making policy exhibits adaptability across distinct new driving scenarios, reaffirming its versatility and robustness.

Further works may focus on the online application of the proposed decision-making policy. Incorporating predictive information for the AEV may enhance its performance. Additionally, exploring the concept of a connected environment to facilitate information sharing among proximate vehicles is worth investigating. The utilization of authentic driving data collected from real-world settings can provide a comprehensive evaluation of the decision-making process within genuine driving conditions.